\documentclass[10pt,twocolumn,letterpaper]{article}

\usepackage[pagenumbers]{cvpr}

\usepackage{iftex}
\ifXeTeX
    \usepackage[T1]{fontenc}
\fi

\usepackage{cuted} 
\usepackage{placeins}

\usepackage{currfile} 

\usepackage{caption} 

\usepackage{algorithm}
\usepackage{algpseudocode}

\definecolor{cvprblue}{rgb}{0.21,0.49,0.74}
\usepackage[breaklinks,colorlinks,allcolors=cvprblue]{hyperref}

\title{When and What to Match in Diffusion-Based Point Tracking}
\newcommand{\correspondingmark}{\raisebox{.45ex}{\fontsize{10}{12}\selectfont\(\dagger\)}}
\author{
  Tianyu Yuan\textsuperscript{1}\qquad
  Yuanbo Yang\textsuperscript{2}\qquad
  Lin-Zhuo Chen\textsuperscript{1}\\[3pt]
  Yao Yao\textsuperscript{1,}\correspondingmark\qquad
  Zhuzhong Qian\textsuperscript{1,}\correspondingmark\\[3pt]
  \textsuperscript{1}Nanjing University\qquad
  \textsuperscript{2}Zhejiang University\\[2pt]
  {\small\correspondingmark\ Corresponding authors}
}
\hypersetup{
  pdftitle={When and What to Match in Diffusion-Based Point Tracking},
  pdfauthor={Tianyu Yuan, Yuanbo Yang, Lin-Zhuo Chen, Yao Yao, Zhuzhong Qian}
}

\begin{document}
\maketitle
\begin{strip}
    \centering
    \includegraphics[width=\textwidth]{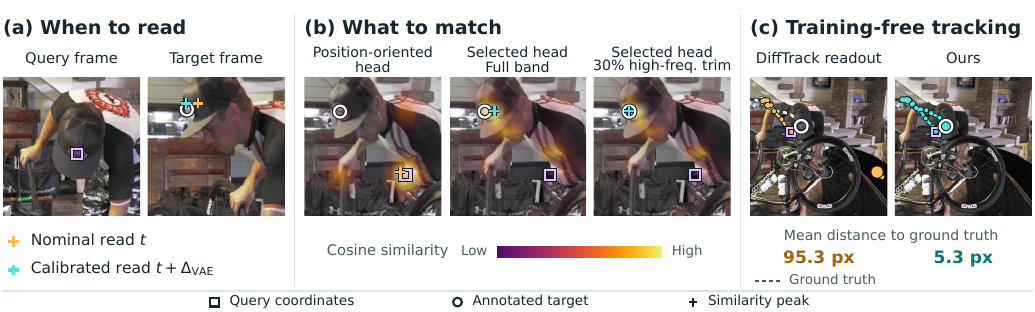}
    \captionsetup{hypcap=false}
    \captionof{figure}{\textbf{When and what to match in diffusion-based point tracking.}
    A TAP-Vid DAVIS example illustrates temporal calibration, head selection, and high-frequency RoPE trimming. The trajectories compare the DiffTrack readout with ours in the same tracker.}
    \label{fig:teaser}
    \captionsetup{hypcap=true}
\end{strip}

\begin{abstract}
Pretrained video diffusion models can support downstream visual perception with spatiotemporal features learned for generation.
Joint denoising encourages these models to relate scene appearance across frames. Point tracking tests how precisely these features localize the same physical point over time.
Controlled translations reveal that video variational autoencoder (VAE) latents can represent earlier scene states than their timestamps suggest.
On a fixed sequence of per-frame features, advancing the readout raises Wan2.1's percentage of correct keypoints (PCK) by 10.2 percentage points on DAVIS.
After temporal calibration, representative attention heads favor the corresponding point, the target object region, or the original query coordinates.
In every layer of the two backbones examined, at least one head outperforms the whole-layer descriptor.
Trimming high-frequency positional channels reduces accuracy near the query and improves it at larger displacements.
These findings guide a training-free tracker that matches calibrated single-head keys with displacement-dependent channel trimming.
Across three backbones and three benchmarks, this readout improves PCK by \mbox{10.2--28.0} percentage points over the matched baseline.
\end{abstract}

\section{Introduction}
\label{sec:intro}

Jointly denoising video frames encourages diffusion models to capture appearance changes caused by object and camera motion~\cite{ho2022videodiffusion,wan2025wan}.
Pretrained video diffusion models support temporally consistent depth estimation~\cite{hu2025depthcrafter,shao2025chronodepth}, and query--key similarities in diffusion transformers (DiTs) capture cross-frame correspondence~\cite{nam2025difftrack}.
Point tracking tests how precisely these features identify the same physical point and estimate its visibility across frames~\cite{doersch2022tap}.
This requires knowing when a feature represents the scene and which location its similarity favors.

Latent frames from a video variational autoencoder (VAE) can represent the scene at an earlier time than their nominal timestamps indicate.
We measure this temporal lag by registering the latents of static and translated clips and find that it varies across the tested encoders.
Multi-phase extraction interleaves features from shifted encodings to cover every input frame while preserving the VAE's native temporal stride in each pass.
With these maps held fixed, shifting the readout by the measured lag improves localization for all three tested backbones.

At calibrated readout times, attention heads vary in the spatial relationships they match.
Representative heads peak near the corresponding point, respond broadly across the target object, or favor the original query coordinates.
A whole-layer descriptor combines these matching preferences in a single similarity score.
On TAP-Vid DAVIS, every layer of Wan2.1 and HunyuanVideo 1.5 contains at least one head whose keys localize points more accurately than the corresponding whole-layer descriptor.

The preference for the original query coordinates motivates examining how positional encoding affects matching.
In rotary position embeddings (RoPE)~\cite{su2021roformer}, a given displacement produces a larger change in relative phase for higher-frequency channel pairs.
On TAP-Vid DAVIS, a lower-frequency band gives higher localization accuracy than an equally sized higher-frequency band for all 20 Wan2.1 heads in our comparison.
Grouping tracking results by displacement shows that retaining high frequencies helps at small displacements and that trimming them becomes beneficial as points move farther from their queries.

Our training-free tracker configures each frozen backbone once using synthetic Kubric videos~\cite{greff2022kubric}.
Controlled translations provide the VAE lag estimate, and annotated tracks guide the choice of a single attention head and the frequency band for each displacement range.
The tracker interpolates the selected head's interleaved keys at the lag-adjusted time and chooses a band using the distance from the original query to the previous prediction.

We evaluate three diffusion backbones on TAP-Vid DAVIS, a 30-video TAP-Vid Kinetics subset, and ITTO-MOSE~\cite{demler2025itto}.
With a shared tracker and matched backbones, our readout improves Average Jaccard (AJ) and percentage of correct keypoints (PCK) over DiffTrack-style features in all nine comparisons, with PCK gains of 10.2--28.0 percentage points.
In the DAVIS ablation, choosing the band by displacement also raises PCK relative to the best tested static trim for both backbones.

\begin{itemize}[label=$\bullet$]
    \item \textbf{Temporal lag in video latents.}
    We measure VAE lag with controlled translations and isolate its effect on localization.
    \item \textbf{What diffusion features match.}
    We characterize head matching behavior and show how the effect of channel trimming changes with displacement.
    \item \textbf{Training-free point tracking.}
    We match temporally calibrated single-head keys with displacement-dependent trimming.
\end{itemize}

\section{Related Work}
\label{sec:related_work}

\paragraph{Point Tracking.}
TAP-Vid benchmarks the localization and visibility of queried points over time~\cite{doersch2022tap}.
PIPs models long-range trajectories through occlusion~\cite{harley2022pips}, and TAPIR initializes matches frame by frame and refines them over time~\cite{doersch2023tapir}.
CoTracker models dependencies between tracks~\cite{karaev2024cotracker,karaev2025cotracker3}, Track-On maintains memory for online tracking~\cite{aydemir2025trackon}, and TAPNext formulates tracking as next-token prediction~\cite{zholus2025tapnext}.
PointOdyssey provides synthetic trajectories for long-term supervision~\cite{zheng2023point}.
Contrastive random walks enable self-supervised tracking from unlabeled videos~\cite{shrivastava2024randomwalks}.
Frozen visual foundation models have also been evaluated as point tracking descriptors~\cite{aydemir2024foundation}.
We address the same task with a frozen video diffusion backbone and a training-free tracker.

\paragraph{Correspondence from Generative Models.}
DIFT extracts features from image diffusion models to match points across images~\cite{tang2023dift}.
Semantic correspondence has also been obtained from diffusion attention~\cite{hedlin2023correspondence} and from combining Stable Diffusion and DINO features~\cite{zhang2023diffusiondino}.
For video generators, KL-tracing estimates optical flow by measuring how a local perturbation changes the distribution of predictions~\cite{kim2025kltracing}.
Point Prompting tracks points by propagating visible query markers through video generation~\cite{shrivastava2026pointprompting}.
DiffTrack studies cross-frame matching in video diffusion transformers (DiTs) across feature types, layers, and denoising steps~\cite{nam2025difftrack}.
DiTracker combines matching costs from a DiT and a ResNet and adapts the DiT with LoRA for learned trajectory refinement~\cite{son2026ditracker}.
Our analysis examines what individual attention heads match within a frozen DiT and uses the keys of a single head as tracking descriptors.

\paragraph{Temporal Sampling and Alignment.}
Latent diffusion operates on compressed visual representations~\cite{rombach2022ldm}, and DiTs apply a transformer to denoise latent patches~\cite{peebles2023dit}.
Video generators such as Wan and CogVideoX compress both space and time with video VAEs~\cite{wan2025wan,yang2025cogvideox,wu2025videovae}.
DiffTrack and the zero-shot evaluation in DiTracker encode each frame independently to obtain one latent per input frame~\cite{nam2025difftrack,son2026ditracker}.
TrackCraft3R assigns matching temporal RoPE indices to geometry and track tokens for dense 3D tracking~\cite{nam2026trackcraft3r}.
Our multi-phase extraction covers each input timestamp by interleaving features from shifted clips, preserving the VAE's native temporal stride within each pass.
The interleaving follows the sampling principle of shift-and-stitch in strided networks~\cite{long2015fcn}.
We measure the lag of the latents against the input video and use it to calibrate the readout time on this sampling grid.

\section{When to Read Features}
\label{sec:temporal_calibration}

Tracking a moving point requires key maps that cover the input frames and align with them in time.
We use multi-phase extraction to obtain a key map for every frame and shift the readout using the temporal lag measured in the video latents.

\subsection{Temporal Lag in VAE Latents}
\label{sec:vae_lag}

Video diffusion models commonly use a variational autoencoder (VAE) to encode videos into sequences of latent frames~\cite{wan2025wan,yang2025cogvideox,wu2025videovae}.
We observe that a latent frame can reflect the scene at an earlier time than its nominal timestamp indicates.
For temporal downsampling factor $r$, latent frame $j$ has \emph{nominal timestamp} $t_j=rj$, with $t_0=0$.
Its \emph{effective timestamp} $\hat t_j$ is the input time at which the scene best matches the latent frame's spatial content.
We define the \emph{temporal lag} as $\Delta_j=t_j-\hat t_j$, measured in input frames.

To estimate $\hat t_j$, we encode a static clip and constant-velocity translations of each synthetic image with the same VAE.
Let $Z_j^0$ and $Z_j^{\mathbf v}$ denote the static and translated latent maps at index $j$, where $\mathbf v$ is the translation velocity in pixels per input frame.
At a candidate input time $s$ in the search set $\mathcal S$, a point at $\mathbf x$ in the static clip appears at $\mathbf x+s\mathbf v$ in the translated clip.
We sample both maps bilinearly at these corresponding positions for each $\mathbf x$ on an interior grid $\Omega$ of image coordinates and estimate the effective timestamp as
\begin{equation}
    \hat t_j = \operatorname*{arg\,min}_{s\in\mathcal S}
    \sum_{\mathbf x\in\Omega}
    \big\|Z_j^{\mathbf v}(\mathbf x+s\mathbf v)-Z_j^0(\mathbf x)\big\|_2^2.
    \label{eq:latent_registration}
\end{equation}
We report the median lag across images, translation directions, and non-initial latent frames.

\subsection{Temporal Calibration of DiT Features}
\label{sec:dit_calibration}

A single VAE pass yields latent maps at nominal timestamps $0,r,2r,\ldots$.
\emph{Multi-phase feature extraction} covers the intervening timestamps while retaining the native stride $r$ in each pass.
For phase $p\in\{0,\ldots,r-1\}$, we prepend $(-p)\bmod r$ copies of the first frame before encoding.
In the original clip's time coordinates, the non-initial latents of phase $p$ have nominal timestamps $t\equiv p\pmod r$. All phases share frame zero.

\begin{figure}[t]
    \centering
    \includegraphics[width=\columnwidth]{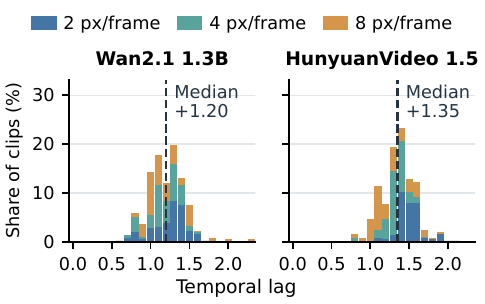}
    \caption{\textbf{VAE temporal lag under synthetic translation.}
    Each clip contributes its median lag over non-initial frames, measured in input frames.
    Bars are stacked by translation speed.
    Each dashed line marks the median across clips for that encoder.}
    \label{fig:vae_temporal_offsets}
\end{figure}

To estimate temporal lag on the dense grid, we interleave the VAE latents by nominal timestamp and apply the registration in \cref{sec:vae_lag} to the synthetic translations.
Each clip contributes the median lag over its non-initial frames.
Across translation directions and speeds, the clip-level estimates cluster around a characteristic lag for each encoder (\cref{fig:vae_temporal_offsets}).
We therefore use the median across clips as the representative lag $\Delta_{\mathrm{VAE}}$.

For tracking, we process each phase with the same frozen diffusion transformer (DiT) and read keys from the selected attention head.
We interleave the keys by nominal timestamp to obtain the key maps $K_0,\ldots,K_{T-1}$ for a $T$-frame clip, averaging the phase maps at frame zero.
We then apply the lag to the interleaved sequence. The calibrated key map for input frame $t$ is
\begin{equation}
    \widetilde K_t = K(t+\Delta_{\mathrm{VAE}}), \qquad 1\leq t<T,
    \label{eq:dit_temporal_calibration}
\end{equation}
where $K(\cdot)$ linearly interpolates $K_a$ and $K_{a+1}$ when the readout time falls between nominal timestamps $a$ and $a+1$.
Readout times outside the clip use the nearest endpoint, and the first frame keeps its nominal map, $\widetilde K_0=K_0$.

\section{What Features to Match}
\label{sec:feature_selection}

We compare attention feature types and evaluate single-head and whole-layer descriptors using the same temporal calibration.
Some heads favor the original query coordinates after the point moves. This motivates analyzing how the RoPE frequencies retained in the calibrated keys affect matching at different displacements.

\begin{figure*}[t]
    \centering
    \includegraphics[width=\textwidth]{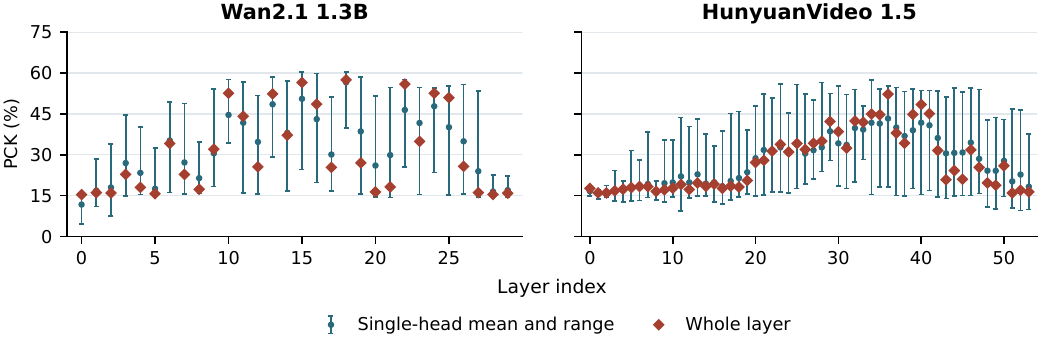}
    \caption{\textbf{Single heads and whole layers on TAP-Vid DAVIS.}
    For each layer, the dot shows the mean PCK across heads and the bar spans the lowest to highest head score.
    The diamond shows the PCK obtained by concatenating all keys in the layer.}
    \label{fig:head_vs_layer_pck}
\end{figure*}

\begin{figure*}[t]
    \centering
    \includegraphics[width=\textwidth]{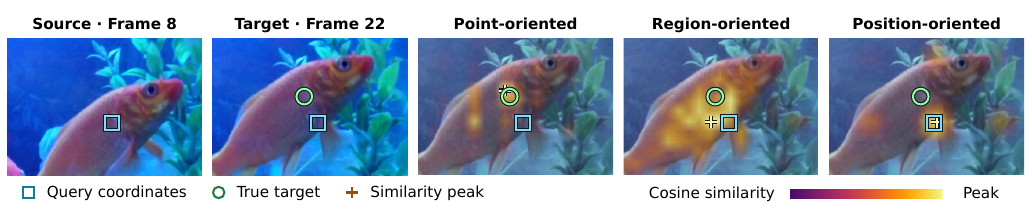}
    \caption{\textbf{Cross-frame key similarity.}
    The three Wan2.1 heads compare the same query point with locations in the same target frame.
    The square marks the query's original coordinates, the circle marks the annotated target, and the cross marks each map's global peak.
    Brighter areas in each overlay indicate higher cosine similarity.}
    \label{fig:head_response_patterns}
\end{figure*}

\subsection{Single Heads vs. Whole Layers}
\label{sec:head_vs_layer}

Each self-attention head provides queries (Q), keys (K), values (V), and outputs (H), any of which can serve as point descriptors.
A descriptor pair specifies which features represent the query point and candidate target locations, respectively.

To compare these choices, we evaluate five pairs across all heads of Wan2.1-1.3B~\cite{wan2025wan} on 30 Kubric videos~\cite{greff2022kubric}, using temporally calibrated multi-phase features throughout.
K--K achieves the highest mean percentage of correct keypoints (PCK) across videos and heads, followed closely by Q--Q (\cref{tab:kubric_descriptor_comparison}).
This result motivates using keys on both sides of the match in the following analyses.

\begin{table}[b]
    \centering
    \caption{\textbf{Descriptor pairs on synthetic Kubric videos.}
    PCK is averaged over all 360 heads of Wan2.1-1.3B.}
    \label{tab:kubric_descriptor_comparison}
    \begin{tabular}{@{}lccccc@{}}
        \toprule
        Descriptor pair & K--K & Q--Q & V--V & H--H & Q--K \\
        \midrule
        Mean PCK (\%) & \textbf{49.38} & 49.19 & 43.50 & 41.47 & 37.38 \\
        \bottomrule
    \end{tabular}
\end{table}

The choice of keys leaves open how many heads should contribute to a descriptor.
For each layer, we compare cosine matching with the keys of a single head and with the concatenated keys of all its heads.
The comparison uses TAP-Vid DAVIS~\cite{doersch2022tap} with Wan2.1-1.3B and HunyuanVideo 1.5~\cite{hunyuanvideo15}.
Both use temporally calibrated multi-phase features and the same tracker settings. \Cref{fig:head_vs_layer_pck} reports PCK averaged over videos.

Although the whole-layer descriptor contains richer feature information, at least one head in every layer of both models achieves higher PCK on its own.
We therefore use a single head's keys as the tracking descriptor.
PCK also varies widely among heads within the same layer.
This spread motivates a closer look at what different heads match.

\subsection{What Different Heads Match}
\label{sec:head_response_patterns}

\begin{figure*}[t]
    \centering
    \includegraphics[width=\textwidth]{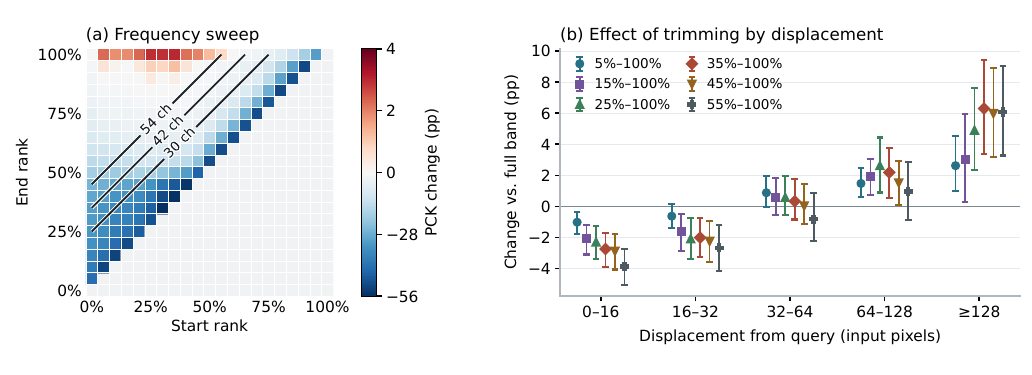}
    \caption{\textbf{Tracking accuracy across RoPE bands and point displacements.}
    Both panels report PCK changes from the full-band baseline, in percentage points, for the same Wan2.1 head on TAP-Vid DAVIS.
    \textbf{(a)} Each cell represents a retained band between two frequency ranks, ordered from the highest frequencies at 0\% to the lowest at 100\%. The diagonals trace fixed rank spans, with channel counts labeled.
    \textbf{(b)} For each of six high-frequency trims, predictions from full-video tracking are grouped by ground-truth displacement from the query location. Error bars show 95\% confidence intervals from paired bootstrap resampling over videos.}
    \label{fig:frequency_trimming_pck}
\end{figure*}

We compare the similarity maps of different heads for the same query point and frame pair.
Each map gives the cosine similarity between the source key, sampled at the query point by bilinear interpolation, and each key on the target feature grid.
The analysis uses temporally calibrated multi-phase keys from Wan2.1.

\Cref{fig:head_response_patterns} illustrates three representative matching behaviors.
The point-oriented head peaks near the annotated target, the region-oriented head responds broadly across the fish, and the position-oriented head peaks near the query's original image coordinates.

We quantify these matching behaviors on 30 Kubric videos with point trajectories and instance masks.
Each sample contains a point visible in two frames of the same clip, separated by 4, 8, 12, 16, or 20 frames.
Among the 30,768 samples with point displacements of at least 64 pixels, the point-oriented and region-oriented heads peak within 16 pixels of the annotated target in $88.0\%$ and $58.4\%$ of cases, respectively.
The position-oriented head peaks within 16 pixels of the original query coordinates in every such sample.

To quantify how broadly each head responds across the target instance, we measure the fraction of patches inside its mask that have strong responses.
A patch is included if its center lies inside the mask. A response is strong if its score lies in the upper $20\%$ of the interval from the map's median to its maximum.
Across the samples with at least eight patches inside the mask, mean coverage is $68.6\%$ for the region-oriented head, $18.9\%$ for the point-oriented head, and $0.6\%$ for the position-oriented head.

For the query in \cref{fig:head_response_patterns}, most layers contain both a head that peaks near the corresponding point and a head that peaks near the original query coordinates.
Matching concatenated keys combines these different preferences in one similarity score, which helps explain why a selected head can track the point more accurately.

\subsection{Channel Selection by RoPE Frequency}
\label{sec:rope_frequency_trimming}

\begin{figure*}[!t]
    \centering
    \includegraphics[width=\textwidth]{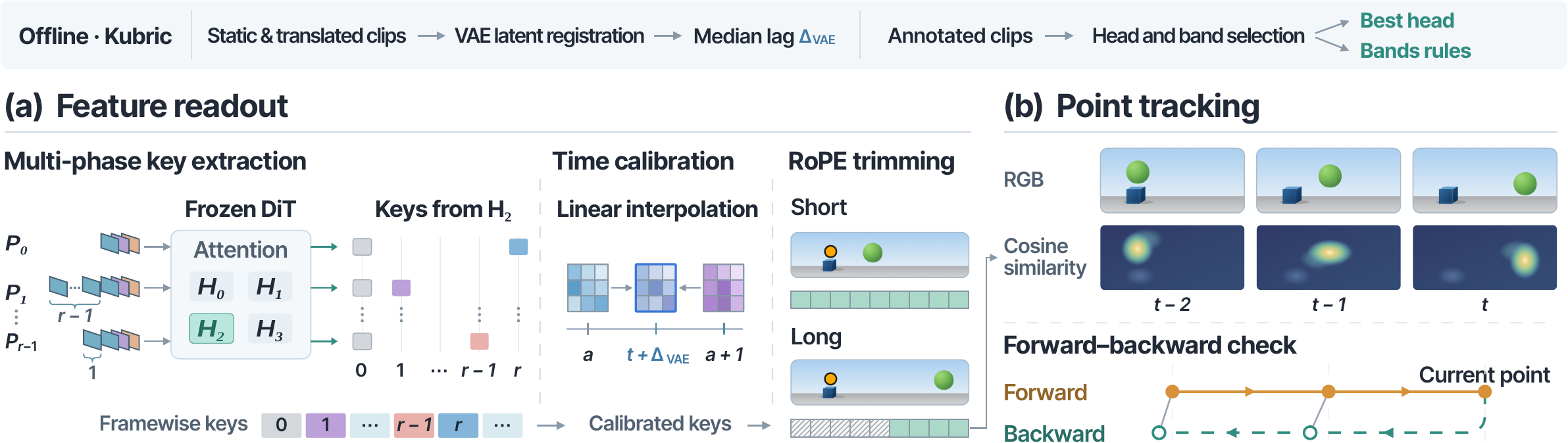}
    \caption{\textbf{Training-free point tracking.}
    We estimate the VAE lag and select the head and RoPE band rule on Kubric. We match temporally calibrated keys over a displacement-dependent band and refine the similarity peak by local soft argmax. Forward--backward consistency gives visibility.}
    \label{fig:point_tracking_pipeline}
\end{figure*}

The position-oriented head in \cref{sec:head_response_patterns} consistently peaks near the original query coordinates after the point moves.
This behavior motivates examining how positional encoding affects cross-frame matching.
Modern video diffusion models commonly use 3D rotary position embeddings (3D RoPE) to encode video token positions~\cite{su2021roformer,wan2025wan,hunyuanvideo15}.
Each pair of query or key channels has a fixed RoPE frequency $\omega$ and is assigned to the time, height, or width axis of the token grid.
RoPE rotates the pair by $\omega$ times the token's coordinate along its axis. Two tokens $\delta$ positions apart along that axis therefore have a relative phase shift of $\omega\delta$.
The phase changes faster with displacement in higher-frequency pairs.

We use the model's fixed RoPE frequencies to select channel pairs from the key descriptors.
On each axis, we order the pairs from highest to lowest frequency and trim those outside a chosen rank interval.
The cutoffs are given as fractions of the ranked list. Multiplying each fraction by the number of pairs on that axis and rounding down gives its cutoff index.
For example, $[0.30,1)$ drops the pairs before the 30\% rank cutoff and retains the lower-frequency pairs.
Cosine similarity uses the same retained pairs in the source and target keys.
All comparisons use temporally calibrated multi-phase features from Wan2.1 and the same tracker settings.

We sweep the start and end ranks in steps of five percentage points on TAP-Vid DAVIS.
For the head ranked first by full-band PCKs, the best sampled band trims part of the high-frequency end and improves PCK over the full band (\cref{fig:frequency_trimming_pck}(a)).
Moving up and to the right along a diagonal increases both rank cutoffs by the same amount, shifting the band toward lower frequencies.
On the upper diagonal, for example, the bands $[0,0.45)$ and $[0.50,0.95)$ both retain 54 channels.
On all three diagonals, the lowest-frequency band with the labeled channel count achieves higher PCK than the band of the same size at the left endpoint.

To test whether the low-frequency advantage extends beyond the top-ranked head, we repeat the equal-channel comparison for the top 20 heads in the same PCK ranking.
With the channel count held fixed, the lower-frequency band yields higher PCK for every head tested.

We next examine how the effect of trimming high-frequency channel pairs varies with displacement.
For the same head, we track points with the full band and six trims that remove increasingly more high-frequency pairs.
After tracking, we group predictions on frames after the query where the annotated point is visible.
The displacement is the Euclidean distance in input pixels from the initial query to the annotated point in that frame.
For each bin, we compute PCK per video and average over videos with samples in that bin.
For displacements under 16 pixels, trimming more high-frequency pairs leads to a larger drop in PCK.
As displacement grows, the effect of each trim shifts from a loss to a gain, and all six improve PCK at 128 pixels or more (\cref{fig:frequency_trimming_pck}(b)).

This trend is consistent with the relative rotation $\omega\delta$ in 3D RoPE.
At small displacements, the spatial phase shift is limited, so high-frequency channels can still provide useful matching cues.
At larger displacements, the greater spatial phase shifts in these channels can lower the similarity between corresponding points across frames.
This displacement dependence motivates the trimming rule in \cref{sec:tracking_descriptor_selection}, which retains more high-frequency channels near the query location and trims more at larger displacements.

\section{Training-Free Point Tracking}
\label{sec:point_tracking}

Given a video and query points, our tracker matches temporally calibrated keys from a frozen video diffusion model to predict each point's position and visibility in subsequent frames (\cref{fig:point_tracking_pipeline}).

\subsection{Offline Configuration}
\label{sec:tracking_descriptor_selection}

We render 30 multi-object videos in Kubric~\cite{greff2022kubric}, each with 50 frames, ground-truth point trajectories, and visibility labels.
The scenes vary in object composition, layout, lighting, and viewpoint.
The videos include translations at different speeds, rotations, articulated motion, collisions, occlusions, and camera motion.
For each backbone, we use this synthetic set to estimate the lag, select a tracking head, and set the displacement-dependent RoPE bands before benchmark evaluation (top row of \cref{fig:point_tracking_pipeline}).

We estimate the lag from static and constant-velocity clips made from each video's first frame, using the latent registration in \cref{sec:dit_calibration}.
The lag calibrates the keys for head selection and tracking (\cref{eq:dit_temporal_calibration}).

We run the tracker with each candidate head's calibrated full-band keys and rank the heads by mean PCK across clips.
The highest-ranked head is used for all test videos of that backbone.

Retaining high-frequency channels helps at small displacements, and trimming them can improve matching at larger displacements.
We use the annotated trajectories to choose one RoPE band for each displacement range for the selected head.

\begin{table*}[t]
    \centering
    \caption{\textbf{Point tracking across three benchmarks.}
    Scores are percentages. DiffTrack-style and DINOv3 features use the shared tracker.
    Bold marks the best training-free result in each column.}
    \label{tab:tracking_benchmarks}
    \small
    \setlength{\tabcolsep}{4pt}
    \begin{tabular*}{\textwidth}{@{\extracolsep{\fill}}l*{9}{r}@{}}
        \toprule
        & \multicolumn{3}{c}{TAP-Vid DAVIS} & \multicolumn{3}{c}{TAP-Vid Kinetics (30)} & \multicolumn{3}{c}{ITTO-MOSE} \\
        \cmidrule(lr){2-4}\cmidrule(lr){5-7}\cmidrule(lr){8-10}
        Method & AJ $\uparrow$ & PCK $\uparrow$ & OA $\uparrow$ & AJ $\uparrow$ & PCK $\uparrow$ & OA $\uparrow$ & AJ $\uparrow$ & PCK $\uparrow$ & OA $\uparrow$ \\
        \midrule
        \multicolumn{10}{@{}l}{\emph{Task-trained trackers}} \\
        TAPIR & 57.15 & 69.39 & 86.01 & 51.66 & 64.17 & 83.71 & 33.58 & 48.09 & 75.62 \\
        BootsTAPIR v2 & 60.98 & 73.19 & 88.49 & 55.77 & 68.16 & 86.38 & 36.81 & 53.55 & 76.18 \\
        LocoTrack-Base & 61.54 & 73.64 & 85.94 & 56.73 & 69.86 & 85.60 & 40.49 & 54.76 & 77.93 \\
        TAPTRv2 & 61.56 & 74.50 & 89.27 & 53.64 & 67.91 & 85.54 & 36.04 & 50.89 & 76.42 \\
        CoTracker3 & 62.32 & 75.19 & 89.39 & 56.38 & 69.90 & 85.89 & 41.84 & 55.59 & 79.69 \\
        CoWTracker & 62.38 & 77.62 & 88.74 & 58.55 & 72.15 & 89.74 & 40.01 & 58.73 & 77.70 \\
        BootsTAPNext & 65.29 & 78.24 & 91.30 & 54.84 & 68.60 & 84.25 & 38.05 & 52.16 & 78.27 \\
        \midrule
        \multicolumn{10}{@{}l}{\emph{Training-free configurations}} \\
        DINOv3 ViT-L/16 & 38.57 & 54.74 & 81.94 & 32.45 & 49.90 & 75.03 & 16.67 & 33.89 & 62.32 \\
        DiffTrack readout (Wan2.1-1.3B) & 36.14 & 50.27 & 77.70 & 25.02 & 39.73 & 73.90 & 14.44 & 26.83 & 60.55 \\
        DiffTrack readout (HunyuanVideo 1.5) & 31.86 & 46.53 & 80.14 & 22.35 & 36.28 & 75.32 & 15.36 & 30.11 & 62.56 \\
        DiffTrack readout (Open-Sora 1.3) & 15.57 & 27.11 & 57.66 & 21.37 & 34.95 & 68.58 & 7.54 & 16.76 & 56.29 \\
        \addlinespace[2pt]
        Ours (Wan2.1-1.3B) & \textbf{47.04} & \textbf{65.44} & \textbf{83.83} & \textbf{36.57} & \textbf{55.18} & \textbf{75.88} & \textbf{23.37} & \textbf{43.75} & \textbf{66.10} \\
        Ours (HunyuanVideo 1.5) & 44.40 & 60.78 & 83.67 & 32.42 & 50.25 & 74.34 & 21.47 & 40.33 & 64.49 \\
        Ours (Open-Sora 1.3) & 39.03 & 55.12 & 74.26 & 31.39 & 48.89 & 70.51 & 16.52 & 31.42 & 59.90 \\
        \bottomrule
    \end{tabular*}
\end{table*}

\subsection{Point Tracking Pipeline}
\label{sec:point_tracking_pipeline}

A track starts at the query position $\mathbf{x}_q$ in frame $q$.
Let $\widetilde{K}_t(\mathbf{u})$ denote the bilinear sample of the calibrated key map $\widetilde K_t$ at image coordinate $\mathbf{u}$.
We initialize the point descriptor with $\widetilde{K}_q(\mathbf{x}_q)$ and set $\hat{\mathbf{x}}_q=\mathbf{x}_q$ and $v_q=1$.

Before matching in frame $t>q$, we estimate displacement from the original query using the previous prediction,
\begin{equation}
    d_t = \lVert \hat{\mathbf{x}}_{t-1} - \mathbf{x}_q \rVert_2,
    \label{eq:trimming_displacement}
\end{equation}
where $\hat{\mathbf{x}}_{t-1}$ and $\mathbf{x}_q$ are both in input pixels.
The displacement range containing $d_t$ determines which high-frequency pairs to trim along each RoPE axis.
We retain the same channel pairs in the point descriptor and the calibrated target keys.

We compute the cosine similarity $s_t(\mathbf{u})$ between the point descriptor and $\widetilde{K}_t(\mathbf{u})$ over the selected RoPE band.
Let $\mathcal{N}$ be the disk of candidate pixels around the similarity peak. Local soft argmax gives
\begin{equation}
    \hat{\mathbf{x}}_t =
    \frac{\sum_{\mathbf{u}\in\mathcal{N}} \exp(s_t(\mathbf{u}))\,\mathbf{u}}
         {\sum_{\mathbf{u}\in\mathcal{N}} \exp(s_t(\mathbf{u}))}.
    \label{eq:tracking_localization}
\end{equation}

We estimate visibility by measuring how closely the backward matches recover the recorded forward trajectory.
Starting with the key sampled at $\hat{\mathbf{x}}_t$, we match backward one frame at a time within the tracking window, using the same RoPE band as the forward match at frame $t$.
Let $\check{\mathbf{x}}_s$ denote the retraced position in frame $s$, and let $\mathcal{V}$ collect the earlier frames in the window with a visible forward prediction.
With a tolerance $\tau$ in input pixels, the visibility flag $v_t$ is one if
\begin{equation}
    \sum_{s\in\mathcal{V}} \lVert \check{\mathbf{x}}_s - \hat{\mathbf{x}}_s \rVert_2^2
    \leq \max(1,|\mathcal{V}|)\,\tau^2,
    \label{eq:tracking_visibility}
\end{equation}
and zero otherwise.

\section{Experiments}
\label{sec:experiments}

\begin{figure*}[t]
    \centering
    \includegraphics[width=\textwidth]{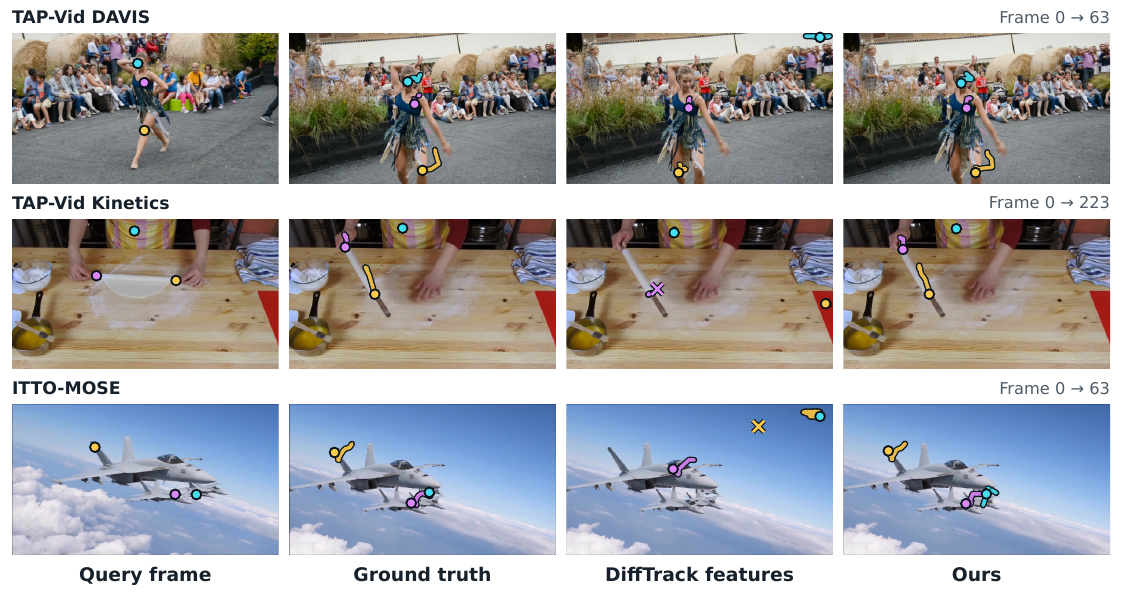}
    \caption{\textbf{Qualitative results on three benchmarks.}
    Both readouts share Wan2.1-1.3B and tracker settings.
    In the target frame, colors distinguish queries, trails show visible segments of the last 16 frames, and crosses mark predicted occlusions.}
    \label{fig:qualitative_tracks}
\end{figure*}

\subsection{Experimental Setup}
\label{sec:experimental_setup}

We evaluate all 30 TAP-Vid DAVIS videos~\cite{doersch2022tap,pont2017davis}, 30 TAP-Vid Kinetics videos randomly drawn from 1,144 clips~\cite{carreira2019kinetics700}, and all 44 ITTO-MOSE videos~\cite{demler2025itto,ding2023mose}.
Under the TAP-Vid \emph{first} protocol, we initialize tracks at their first visible annotations and score all subsequent frames through the end of the video.
We average AJ, PCK ($\delta^x_{\mathrm{avg}}$), and occlusion accuracy (OA) over videos. AJ and PCK average over thresholds of $1$, $2$, $4$, $8$, and $16$ pixels.

We extract multi-phase keys at the final denoising step of frozen Wan2.1-1.3B~\cite{wan2025wan}, HunyuanVideo 1.5~\cite{hunyuanvideo15}, and Open-Sora 1.3~\cite{hpcaitech2025opensora13}.
Each backbone uses the same Kubric configuration across benchmarks (\cref{sec:tracking_descriptor_selection}).

Feature readouts share the tracking pipeline. Following DiffTrack~\cite{nam2025difftrack}, the baseline uses framewise VAE encoding and full-band whole-layer queries and keys from the best layer of each matched backbone.
The DINOv3 baseline uses final-layer patch tokens from a frozen ViT-L/16~\cite{simeoni2025dinov3}.

We also evaluate task-trained trackers using released checkpoints, including TAPIR~\cite{doersch2023tapir}, BootsTAPIR~\cite{doersch2024bootstap}, LocoTrack~\cite{cho2024locotrack}, TAPTRv2~\cite{li2024taptrv2}, CoTracker3~\cite{karaev2025cotracker3}, CoWTracker~\cite{lai2026cowtracker}, and BootsTAPNext~\cite{zholus2025tapnext}.

\subsection{Point Tracking Results}
\label{sec:tracking_results}

\Cref{tab:tracking_benchmarks} compares our features with the DiffTrack-style readout using the same diffusion backbones and tracking pipeline.
Our features improve AJ and PCK in all nine backbone--benchmark comparisons, with PCK gains of 10.2--28.0 percentage points.
The gains extend to Kinetics and ITTO-MOSE with the same Kubric-selected configurations.

Among the training-free configurations, our Wan2.1 variant achieves the highest AJ, PCK, and OA on all three benchmarks.
Compared with DINOv3 features in the shared tracker, its PCK gains range from 5.3 to 10.7 percentage points.
Task-trained trackers achieve higher AJ, PCK, and OA across the three benchmarks.

\Cref{fig:qualitative_tracks} shows that our method produces more accurate and stable point tracks than the DiffTrack-style baseline.

\subsection{Component Ablation}
\label{sec:component_ablation}

\Cref{tab:component_ablation} examines temporal calibration, head selection, and frequency trimming on DAVIS for Wan2.1 and HunyuanVideo.
Every variant uses multi-phase keys and the same tracker settings.
For each backbone, both whole-layer variants use the best layer as measured by calibrated full-band PCK.
Advancing the readout at that layer increases PCK by 8.3 percentage points on Wan2.1 and 2.6 points on HunyuanVideo.
Switching to the selected head adds 2.8 and 5.2 points, respectively.

The calibrated single head provides a common starting point for comparing frequency trims.
For the static reference, we sweep cutoffs from 5\% to 50\% in five-percentage-point steps and select the one with the highest PCK.
The displacement-dependent bands selected on Kubric yield the highest PCK for both backbones, gaining 2.63 points on Wan2.1 and 2.18 points on HunyuanVideo over the best static trim.

\begin{table}[t]
    \centering
    \caption{\textbf{Component ablation on TAP-Vid DAVIS.}
    Wan2.1 uses layer L18 and head L18H11 with a 1.2-frame advance. HunyuanVideo uses layer L36 and head L34H8 with a 1.35-frame advance.
    The static cutoffs are 35\% and 45\% from the high-frequency end, respectively.}
    \label{tab:component_ablation}
    \small
    \setlength{\tabcolsep}{2.5pt}
    \begin{tabular*}{\columnwidth}{@{\extracolsep{\fill}}lllrr@{}}
        \toprule
        & & & \multicolumn{2}{c}{PCK (\%) $\uparrow$} \\
        \cmidrule(l){4-5}
        Readout & Descriptor & RoPE band & Wan2.1 & HunyuanVideo \\
        \midrule
        Nominal & Whole layer & Full band & 49.20 & 49.67 \\
        Calibrated & Whole layer & Full band & 57.47 & 52.22 \\
        Calibrated & Single head & Full band & 60.31 & 57.46 \\
        Calibrated & Single head & Static trim & 62.81 & 58.60 \\
        Calibrated & Single head & Dynamic trim & \textbf{65.44} & \textbf{60.78} \\
        \bottomrule
    \end{tabular*}
\end{table}

\subsection{Temporal Sampling and Calibration}
\label{sec:temporal_sampling}

\begin{table}[t]
    \centering
    \caption{\textbf{Temporal sampling and calibration on TAP-Vid DAVIS.}
    Macro PCK (\%) with the same head across readouts for each backbone.}
    \label{tab:temporal_readouts}
    \small
    \begin{tabular*}{\columnwidth}{@{\extracolsep{\fill}}lrrr@{}}
        \toprule
        Readout & Wan2.1 & Hunyuan & Open-Sora \\
        \midrule
        Framewise & 56.32 & 51.27 & 44.09 \\
        Stride 1 & 57.88 & 53.82 & 40.05 \\
        Pre-DiT upsampling & 46.56 & 45.71 & 32.87 \\
        Post-DiT upsampling & 47.46 & 49.62 & 37.23 \\
        Multi-phase, nominal & 50.12 & 55.68 & 41.93 \\
        Multi-phase, calibrated & \textbf{60.31} & \textbf{57.46} & \textbf{47.15} \\
        \bottomrule
    \end{tabular*}
\end{table}

\Cref{tab:temporal_readouts} compares six readouts on full TAP-Vid DAVIS videos.
Framewise encoding runs the VAE separately for each frame and the DiT jointly. Stride 1 encodes the clip jointly with VAE temporal downsampling disabled.
The upsampling variants retain the native stride and interpolate VAE latents before the DiT or keys after it.
Both multi-phase variants use the same interleaved keys from \cref{sec:dit_calibration}.

At nominal times, multi-phase extraction outperforms both upsampling variants on all three backbones.
Calibrating the same keys yields the highest PCK, gaining 10.2 points on Wan2.1, 1.8 on HunyuanVideo, and 5.2 on Open-Sora.

\section{Conclusion}
\label{sec:conclusion}

Our training-free tracker improves point localization across three benchmarks by correcting latent lag and matching a selected head's keys with displacement-dependent trimming.
Shifting the readout of a fixed sequence of per-frame features improves localization, showing that temporal alignment matters even when every input frame has a feature map.
The point- and region-oriented responses suggest that a single frozen video diffusion model could serve perception tasks at different spatial granularities.

\section*{LLM Usage Statement}
We used OpenAI Codex to assist with revising the manuscript text and refining the implementation code.
It also assisted with experiment design, execution, and analysis of results, as well as with revising figures and diagrams.
The authors take full responsibility for the contents of this manuscript.

\clearpage
\maketitlesupplementary
\appendix
\numberwithin{equation}{section}
\setcounter{figure}{0}
\setcounter{table}{0}
\renewcommand{\thefigure}{S\arabic{figure}}
\renewcommand{\thetable}{S\arabic{table}}
\renewcommand{\thealgorithm}{S\arabic{algorithm}}
\renewcommand{\theHfigure}{supp.\arabic{figure}}
\renewcommand{\theHtable}{supp.\arabic{table}}
This supplement provides synthetic-data and extraction details, an overall tracking algorithm, and additional experiments.
The experiments examine head combinations, qualitative results, resource costs, and denoising-step selection.
Section numbers without an appendix prefix refer to the main paper.

\section{Synthetic Calibration Data}
\label{sec:supp_data}

\subsection{Scene and Motion Diversity}

The offline configuration uses 30 independently rendered Kubric videos~\cite{greff2022kubric}.
Each video contains 50 RGB frames at $832\times480$ resolution and 24 frames per second.
The scenes contain 12--17 scanned object instances drawn from 50 distinct assets, with backgrounds sampled from eight environment maps.
Scene seeds control the object selection, size, placement, orientation, lighting, and camera viewpoint.
This variation produces cluttered scenes with different object scales, appearances, and occlusion patterns.

\begin{figure*}[t]
    \centering
    \includegraphics[width=\textwidth]{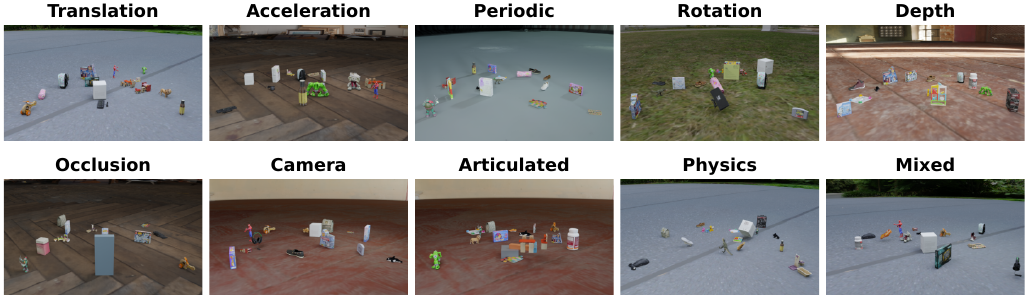}
    \caption{\textbf{Examples from the synthetic calibration set.}
    A frame from one video in each motion family illustrates the variation in objects, scene layout, lighting, and viewpoint.
    Motion families and their temporal variations are listed in \cref{tab:supp_motions}.}
    \label{fig:supp_scenes}
\end{figure*}

The videos cover ten motion families, with three variants per family (\cref{tab:supp_motions}).
The assigned family specifies the dominant motion, while other objects provide background clutter and potential occluders.
Camera motion, changes in depth, articulated motion, and physical interactions complement the image-plane translations used in the controlled lag measurement.
\Cref{fig:supp_scenes} shows one scene from each family.

\begin{table}[b]
    \centering
    \caption{\textbf{Motion coverage of the calibration videos.}
    Each row contributes three videos with independent scene seeds.}
    \label{tab:supp_motions}
    \small
    \begin{tabular}{@{}lp{0.64\columnwidth}@{}}
        \toprule
        Family & Variants \\
        \midrule
        Translation & Slow, medium, and fast \\
        Acceleration & Ease-in, stop-and-go, and reversal \\
        Periodic & Orbit, zigzag, and rapid oscillation \\
        Rotation & Yaw, rolling, and tumbling \\
        Depth & Approach, recession, and depth with zoom \\
        Occlusion & Crossing, foreground occlusion, and reentry \\
        Camera & Truck with pan, orbit, and dolly with roll \\
        Articulated & Single joint, two joints, and fast linkage \\
        Physics & Drop, bounce with roll, and collision \\
        Mixed & Combined object and camera motions \\
        \bottomrule
    \end{tabular}
\end{table}

\subsection{Trajectories, Masks, and Calibration Inputs}

Each video provides 96 surface-point trajectories, giving 2,880 trajectories across the set.
A sampled surface point is stored in its object's local 3D coordinates and projected through the object and camera transforms in each frame.
Visibility requires an unobstructed camera ray and agreement with the rendered instance mask near the projection.
Occlusion, self-occlusion, and leaving the image make the point invisible.
Each trajectory has at least eight visible frames.
Rendering uses the instantaneous scene geometry with motion blur disabled.

The same scenes support three distinct offline decisions.
Static images taken from their first frames generate the controlled translations for estimating VAE lag.
The rendered videos and trajectories determine the tracking head.
The trajectories also provide the displacement groups used to configure dynamic trim.
The resulting lag, head, and trim rule are fixed for each backbone before benchmark evaluation.

\section{Temporal Extraction and Calibration}
\label{sec:supp_temporal}

\subsection{Controlled Measurement of VAE Lag}

Known image translations provide a common spatial reference for measuring time in the latent maps.
For each of the 30 synthetic images, we construct a 25-frame static clip and translated clips moving left, right, up, or down at 2, 4, or 8 pixels per frame.
This gives 360 moving clips per encoder.
Translation uses reflection at the image boundary to fill the newly exposed area.
Registration samples the interior shared by the static image and translated images, reducing the influence of the boundary fill.

The VAE posterior mode is used for both static and moving clips.
Multi-phase encoding places their latent maps on the dense input-frame timeline, as described below.
For a map with nominal timestamp $t$, let $Z_t^0$ and $Z_t^{\mathbf v}$ denote the static and moving latents, respectively.
The candidate effective time is found by minimizing
\begin{equation}
    E_t(s)=\frac{1}{|\Omega|D_z}
    \sum_{\mathbf x\in\Omega}
    \left\|Z_t^{\mathbf v}(\mathbf x+s\mathbf v)
    -Z_t^0(\mathbf x)\right\|_2^2,
    \label{eq:supp_registration}
\end{equation}
where $D_z$ is the latent channel count and $\Omega$ is the shared interior sampling grid.
Coordinates are expressed in input pixels and mapped to the latent grid for bilinear sampling.
The same spatial support is used for every candidate time of a given motion.
Thus, a candidate cannot obtain a smaller error by discarding difficult locations.

Candidate times are sampled every 0.1 input frame within $[0,24]$ and within four frames of the nominal timestamp.
An additional search with an eight-frame radius checks whether the narrower search truncates the estimate.
The interior grid has up to $16\times16$ sample locations, adjusted to the common overlap for large translations.
These searches operate on the encoded maps, so changing the time-search grid does not require another VAE pass.

For each clip, the lag is the median of $t-\arg\min_s E_t(s)$ over its non-initial maps.
The encoder's representative lag is the median of the clip-level estimates.
This aggregation reduces the influence of individual registration failures and gives each image--motion combination the same weight.
The median of an even number of estimates can lie halfway between search-grid values, so the aggregate need not be an integer multiple of 0.1 frame.
The measurement is made on VAE latents before the DiT and does not depend on an attention head.

\subsection{Multi-Phase Time Coordinates}

Multi-phase extraction supplies denser temporal samples while preserving the VAE's native temporal stride in every pass.
Let $r$ be that stride and let phase $p$ prepend
\begin{equation}
    b_p=(-p)\bmod r
\end{equation}
copies of the first input frame.
For non-initial latent index $j\geq1$, the nominal input time is
\begin{equation}
    t_{p,j}=rj-b_p.
    \label{eq:supp_phase_time}
\end{equation}
Only maps whose nominal times lie in the original clip are retained.
The initial map of every phase is assigned to input frame zero.

\Cref{tab:supp_phase_times} gives the complete example for $r=4$ and 25 input frames.
Each pass supplies seven maps.
The six non-initial maps from each of four phases cover frames 1--24 exactly once.
The four initial maps are averaged into the map for frame zero, giving 25 maps after interleaving.
The intervening maps come from additional encodings with shifted temporal origins.

\begin{table}[t]
    \centering
    \caption{\textbf{Nominal times in a 25-frame example.}
    A native stride of four gives four phases. Times are indexed from zero in the original clip.}
    \label{tab:supp_phase_times}
    \small
    \begin{tabular*}{\columnwidth}{@{\extracolsep{\fill}}ccl@{}}
        \toprule
        Phase $p$ & Prefix $b_p$ & Retained nominal times \\
        \midrule
        0 & 0 & 0, 4, 8, 12, 16, 20, 24 \\
        1 & 3 & 0, 1, 5, 9, 13, 17, 21 \\
        2 & 2 & 0, 2, 6, 10, 14, 18, 22 \\
        3 & 1 & 0, 3, 7, 11, 15, 19, 23 \\
        \bottomrule
    \end{tabular*}
\end{table}

For tracking, each phase is processed by the same frozen DiT, and the extracted key maps are interleaved using these time coordinates.
Temporal calibration then changes only where this dense sequence is read.
For $t>0$, define
\begin{equation}
    u=\operatorname{clip}(t+\Delta_{\mathrm{VAE}},0,T-1).
\end{equation}
Writing $a=\lfloor u\rfloor$ and $\alpha=u-a$, the calibrated map is
\begin{equation}
    \widetilde K_t=(1-\alpha)K_a+
    \alpha K_{\min(a+1,T-1)}.
    \label{eq:supp_interpolation}
\end{equation}
The first frame keeps $K_0$.
For example, a 1.2-frame lag gives $\widetilde K_9=0.8K_{10}+0.2K_{11}$.
Near the end of a 25-frame window, $\widetilde K_{23}=K_{24}$ because the requested time 24.2 exceeds the available endpoint.
Calibration changes the feature readout time while leaving the input video, model weights, and extracted phase maps fixed.

\section{Descriptors and Head Responses}
\label{sec:supp_descriptors}

\subsection{Feature Definitions and Backbone Settings}

The descriptor comparison uses the per-head queries Q, keys K, values V, and self-attention outputs H.
Queries and keys are recorded after positional rotation, and values are recorded from the value projection.
For head $h$, the output at token $i$ is
\begin{equation}
    H_i^h=\sum_j A_{ij}^h V_j^h,
    \label{eq:supp_hidden}
\end{equation}
where $A^h$ is that head's attention matrix.
H is recorded before concatenating the heads and applying the attention output projection.
This definition keeps its channel dimension equal to that of Q, K, and V for the same head.

For Q--Q, K--K, V--V, and H--H, the source point and target map use the same feature type.
For Q--K, a query vector at the source point is matched to target keys.
All five comparisons use the same extraction times and tracker.
The descriptor-type table in the main paper averages PCK over videos and all 360 heads of Wan2.1, rather than selecting a separate best head for each feature type.

\Cref{tab:supp_backbones} lists the extraction geometry for the three backbones~\cite{wan2025wan,hunyuanvideo15,hpcaitech2025opensora13}.
The VAE and DiT remain frozen, and extraction uses the final denoising step of the configured schedule.
The same selected head is used across benchmarks for each model.
For example, a 25-frame Wan2.1 window gives a single-head tensor of shape $[25,128,30,52]$ after phase interleaving, ordered as time, channel, height, and width.
Calibration preserves this shape, and channel trim changes only its channel dimension.

\begin{table}[t]
    \centering
    \caption{\textbf{Backbone extraction settings.}
    Sizes are height $\times$ width. The pair counts give the time, height, and width components of a single head's RoPE channels.}
    \label{tab:supp_backbones}
    \small
    \setlength{\tabcolsep}{3pt}
    \begin{tabular*}{\columnwidth}{@{\extracolsep{\fill}}lccc@{}}
        \toprule
        & Wan2.1 & Hunyuan & Open-Sora \\
        \midrule
        Input size & $480{\times}832$ & $480{\times}848$ & $720{\times}1280$ \\
        Key grid & $30{\times}52$ & $30{\times}53$ & $45{\times}80$ \\
        Channels/head & 128 & 128 & 72 \\
        RoPE pairs & 22/21/21 & 8/28/28 & 12/12/12 \\
        Temporal stride & 4 & 4 & 4 \\
        Schedule steps & 50 & 50 & 30 \\
        Selected head & L18H11 & L34H8 & L30H10 \\
        \bottomrule
    \end{tabular*}
\end{table}

\subsection{Quantifying Response Patterns}

The response analysis compares heads on identical source points and target frames.
For each visible source--target pair, the source vector is bilinearly sampled at the annotated query coordinate and compared with every target-grid vector using cosine similarity.
The global peak is mapped back to input-image coordinates.
The Kubric analysis uses frame separations of 4, 8, 12, 16, and 20 frames, retaining points visible at both endpoints.
Peak distances are measured separately from the annotated target and the original query coordinates.
These two references distinguish motion-following peaks from peaks that remain at the initial position.

Object coverage measures the spatial extent of a response within the target instance.
Let $s_j$ be the similarity at target patch $j$, $m$ the median similarity over the map, and $M$ its maximum.
The strong-response threshold is
\begin{equation}
    \theta=m+0.8(M-m).
    \label{eq:supp_response_threshold}
\end{equation}
This threshold uses the upper part of the score interval from the median to the maximum.
It does not require a fixed fraction of patches to be strong.
Let $\mathcal I$ be the patches whose centers fall inside the target instance mask.
Coverage is
\begin{equation}
    C=\frac{|\{j\in\mathcal I:s_j\geq\theta\}|}{|\mathcal I|}.
    \label{eq:supp_coverage}
\end{equation}
A patch crossing the object boundary is included according to its center.
We average coverage over samples with at least eight patches in $\mathcal I$, reducing the effect of very small projected instances.

\section{Tracking Details}
\label{sec:supp_tracking}

\subsection{Hyperparameter Configuration}

The offline configuration separates the feature readout from the tracking controls.
The VAE lag is estimated by registration on synthetic translations.
With that lag fixed, heads are ranked by calibrated full-band PCK on the rendered Kubric videos, and retained bands are compared within displacement groups on the same synthetic set.
These three model-dependent choices are fixed before benchmark evaluation.

The localization radius controls the spatial support of soft argmax, while the consistency tolerance controls the allowed backward disagreement.
We select these tracking hyperparameters by grid search, obtaining a localization radius of 17 pixels and a consistency tolerance of 18 pixels.
Both values are fixed across the head-combination and denoising-step studies below, so these comparisons change only the feature readout under study.

The window length and point-batch size are computational settings.
For native temporal stride four, a 25-frame window contains the initial frame and six complete stride intervals, giving seven latent maps in each phase.
A point-batch limit of 128 accommodates the calibration queries without changing their query times or the scoring protocol.
The denoising-step sensitivity is evaluated on synthetic videos in \cref{sec:supp_denoising}, keeping the layer, lag, and tracking controls fixed.

\subsection{Overall Algorithm}

\Cref{alg:supp_tracking} summarizes the offline configuration and test-time pipeline in Section~5.
The feature selection is shared across test videos, while the retained band can change with each point's predicted displacement.
Localization and visibility follow the same rules as the main paper.

\begin{algorithm}[t]
    \caption{Training-free point tracking}
    \label{alg:supp_tracking}
    \small
    \begin{algorithmic}[1]
        \Require Frozen video model, synthetic set $\mathcal S$
        \Statex \textbf{Offline configuration}
        \State Estimate VAE lag $\Delta$ from synthetic translations
        \State Rank calibrated full-band heads by PCK on $\mathcal S$
        \State Select head $h$ and displacement-to-band rule $B$
        \Statex \textbf{Test-time tracking}
        \Require Video $I$, queries $(q_i,\mathbf x_{i,q_i})$, radius $\rho$, tolerance $\tau$
        \For{each extraction window}
            \State Extract head-$h$ keys from all native-stride phases at the selected denoising step
            \State Interleave phases and read at $t+\Delta$ to obtain $\widetilde K_t$
            \For{each frame $t$ in the window}
                \State Initialize arriving descriptors $\mathbf f_i\gets\widetilde K_t(\mathbf x_{i,q_i})$
                \State Set arriving positions to $\mathbf x_{i,q_i}$ and visibility to one
                \For{each query $i$ with $q_i<t$}
                    \State $d\gets\|\hat{\mathbf x}_{i,t-1}-\mathbf x_{i,q_i}\|_2$
                    \State Select the same channels $B(d)$ in $\mathbf f_i$ and $\widetilde K_t$
                    \State Compute their cosine-similarity map $s_{i,t}$
                    \State $\hat{\mathbf x}_{i,t}\gets\operatorname{LocalSoftArgmax}(s_{i,t};\rho)$
                    \State Retrace $\hat{\mathbf x}_{i,t}$ through earlier maps using $B(d)$
                    \State $e\gets\sum_{s\in\mathcal V}\|\check{\mathbf x}_{i,s}-\hat{\mathbf x}_{i,s}\|_2^2$
                    \State $v_{i,t}\gets[e\leq\max(1,|\mathcal V|)\tau^2]$
                \EndFor
            \EndFor
            \State Carry track states and the last real key map forward
        \EndFor
        \State \Return Positions and visibility on original video frames
    \end{algorithmic}
\end{algorithm}

\subsection{Long Videos and Window Boundaries}

Feature extraction divides a video into successive 25-frame windows.
An incomplete final window repeats its last RGB frame to meet the extraction length.
The original frame count is retained, and the repeated frames are excluded from initialization and evaluation.
Temporal interpolation is performed in the current window's coordinates, with endpoint clamping as in \cref{eq:supp_interpolation}.

The tracking state carries the current descriptor, predicted position, visibility, and original query position across windows.
The last real feature map of the preceding window is retained for the backward consistency check at the transition.
For example, after predicting frame 25 in the new window, the first backward step uses the saved map for frame 24 and compares its match with the recorded frame-24 position.
The carried map includes the temporal calibration applied in its own extraction window.

Queries can start at any annotated frame.
A query that first appears in a later window is initialized when that window is reached.
This handling permits full-video evaluation under the first-visible protocol.

\begin{figure}[!t]
    \centering
    \includegraphics[width=\columnwidth]{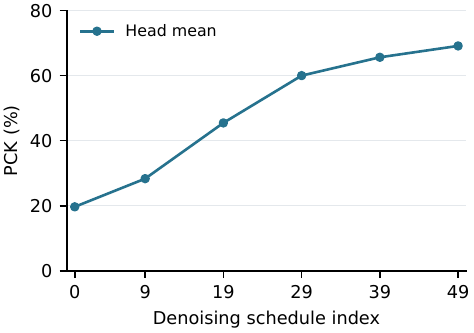}
    \caption{\textbf{Denoising-step sensitivity on Kubric.}
    Each schedule position is evaluated independently with the same clean latents and noise seed.
    The curve shows the mean PCK over all 12 heads in Wan2.1 layer 18.}
    \label{fig:supp_denoising}
\end{figure}

\begin{figure*}[p]
    \centering
    \includegraphics[width=\textwidth]{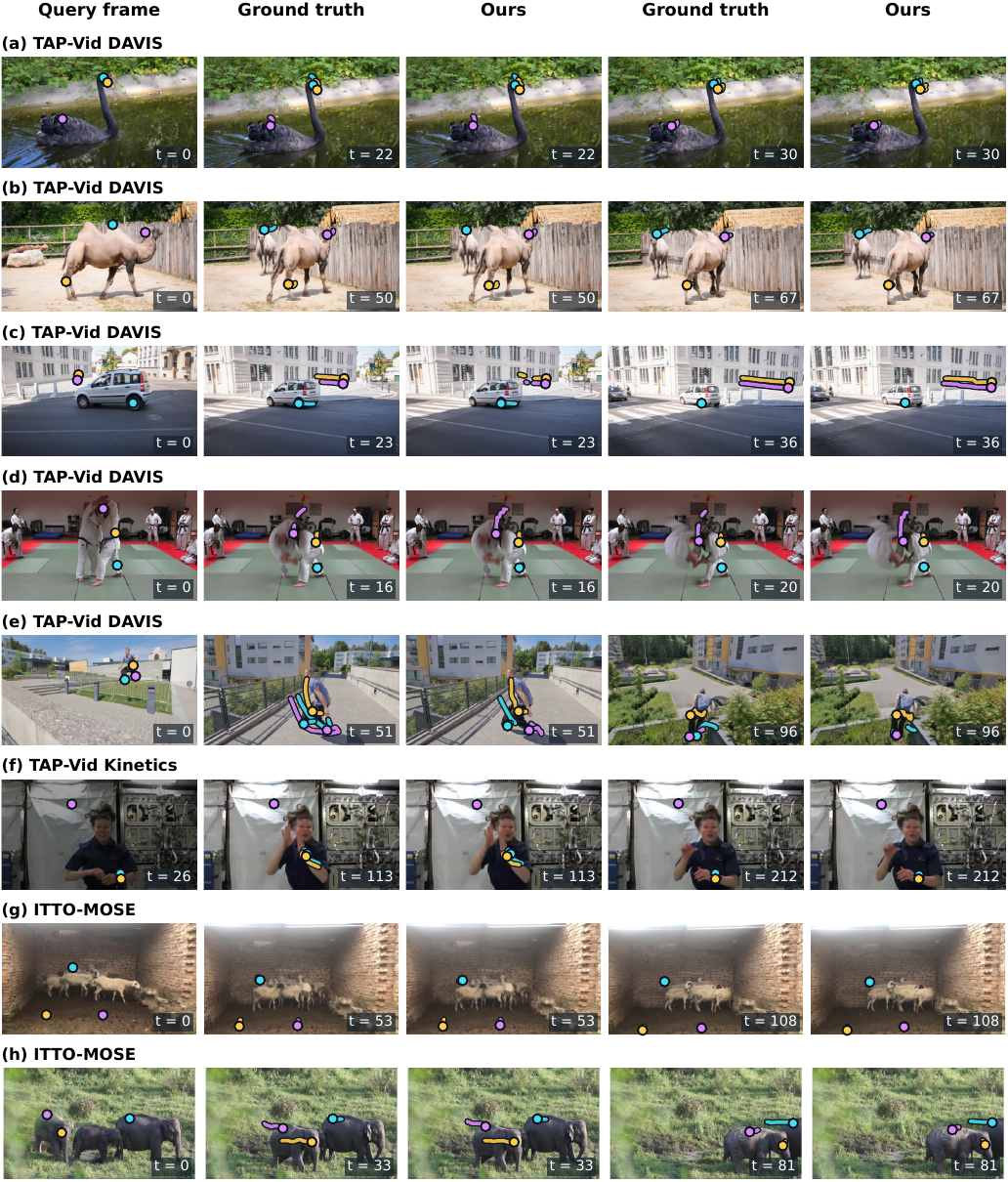}
    \caption{\textbf{Additional full-video tracking examples.}
    Eight sequences from DAVIS, Kinetics, and ITTO-MOSE show our Wan2.1 readout.
    Each row shows the query frame, followed by ground truth and our predictions at two target frames.
    Colors identify the same three points, and trails show their visible positions over the preceding 16 frames.
    These successful examples use video scenes absent from the main-paper figures.}
    \label{fig:supp_qualitative}
\end{figure*}

\section{Top-$k$ Head Combinations}
\label{sec:supp_topk}

We test whether adding high-scoring heads improves the selected single-head descriptor.
Every combination retains the selected Wan2.1 head and adds the highest-ranked remaining heads by calibrated K--K PCK on Kubric.
The keys are concatenated before cosine normalization.
All rows use full-band multi-phase features, a 1.2-frame calibration, and the same tracker on all 30 complete DAVIS videos.
This construction fixes the initial head and changes only the number of additional descriptors.

\begin{table}[t]
    \centering
    \caption{\textbf{Adding high-ranked heads to the selected descriptor.}
    All combinations retain the selected head and add candidates by calibrated Kubric PCK ranking. Scores are percentages on full DAVIS videos.}
    \label{tab:supp_topk}
    \small
    \begin{tabular*}{\columnwidth}{@{\extracolsep{\fill}}rrrrr@{}}
        \toprule
        Heads & Channels & AJ $\uparrow$ & PCK $\uparrow$ & OA $\uparrow$ \\
        \midrule
        1 & 128 & \textbf{42.86} & \textbf{60.31} & \textbf{82.10} \\
        3 & 384 & 42.12 & 59.84 & 81.50 \\
        5 & 640 & 41.90 & 59.72 & 81.62 \\
        10 & 1280 & 42.01 & 59.75 & 81.61 \\
        \bottomrule
    \end{tabular*}
\end{table}

The selected single head achieves the highest AJ, PCK, and OA among the tested combinations (\cref{tab:supp_topk}).
Adding heads lowers PCK by 0.47--0.59 percentage points despite increasing descriptor width.
These results support using the selected head as a compact tracking descriptor.

\section{Time and Memory Costs}
\label{sec:supp_cost}

The resource comparison uses the same Wan2.1 checkpoint and $480\times832$ input resolution, with one RTX A6000 per process.
VAE encoding, DiT execution, and tracking are timed separately, with explicit CUDA synchronization.
The inputs are decoded RGB arrays.
Total time additionally includes preprocessing, latent preparation, temporal interpolation, and feature-cache IO.
The temporal variants use the selected head's K--K descriptors in 25-frame windows, while DiffTrack retains its 13-frame anchored contexts and whole-layer Q--K descriptors.
Timed tracking includes reading that cache and predicting positions and visibility.
Model initialization and one warmup are excluded.
The multi-phase row uses the full band with temporal calibration.

\begin{table}[t]
    \centering
    \caption{\textbf{Single-GPU time and memory on Wan2.1.}
    Times are means over 25-frame prefixes of three DAVIS videos with 23--24 queries, after one warmup, on an RTX A6000.
    Total includes preprocessing and cache IO; GPU memory is peak PyTorch allocation.}
    \label{tab:supp_cost}
    \small
    \setlength{\tabcolsep}{2pt}
    \begin{tabular*}{\columnwidth}{@{\extracolsep{\fill}}lrrrrr@{}}
        \toprule
        Readout & VAE & DiT & Track & Total & GPU \\
        & (s) & (s) & (s) & (s) & (GiB) \\
        \midrule
        Framewise & 1.53 & 2.48 & 0.52 & 5.24 & 5.04 \\
        Stride 1 & 1.50 & 2.49 & 0.51 & 5.19 & 6.41 \\
        Pre-DiT upsampling & 1.27 & 2.50 & 0.51 & 4.98 & 6.42 \\
        Post-DiT upsampling & 1.27 & 0.41 & 0.49 & 2.81 & 6.42 \\
        DiffTrack readout & 1.53 & 1.85 & 1.29 & 5.66 & 8.46 \\
        Multi-phase & 5.10 & 1.63 & 0.50 & 8.02 & 6.55 \\
        \bottomrule
    \end{tabular*}
\end{table}

\begin{samepage}
Multi-phase extraction spends less time in the DiT than the framewise, stride-1, and pre-DiT variants, which process a longer latent sequence jointly.
Its repeated VAE passes nevertheless dominate total runtime.
Compared with the DiffTrack readout, it uses less peak GPU memory but takes longer overall.
\end{samepage}

\section{Denoising-Step Selection}
\label{sec:supp_denoising}

We evaluate independently noised inputs at six positions in a 50-step denoising schedule.
Every position uses the same clean multi-phase VAE latents and noise seed, and only its first DiT forward pass is read.
This avoids confounding the noise level with a history of preceding denoising updates.
The experiment uses all 30 complete 50-frame Kubric videos and all 12 heads in Wan2.1 layer 18, with the lag and tracker fixed.

The schedule index increases from high to low noise.
For the native flow-matching schedule, the tested noise scales are approximately $1.00$, $0.93$, $0.83$, $0.69$, $0.46$, and $0.06$.
At scale $\sigma$, the DiT receives the normalized clean latent $z$ mixed with a fixed noise sample $\epsilon$ as
\begin{equation}
    z_\sigma=(1-\sigma)z+\sigma\epsilon.
\end{equation}
No preceding denoising updates are performed.

Mean PCK over heads increases toward the final step (\cref{fig:supp_denoising}).
The final tested step yields a mean PCK of 69.11\%.
This supports the final-step readout for Wan2.1 under the synthetic calibration protocol.

\section{More Qualitative Results}
\label{sec:supp_qualitative}

\Cref{fig:supp_qualitative} shows additional successful tracking examples from DAVIS, Kinetics, and ITTO-MOSE.
Each row compares our Wan2.1 predictions with ground truth at two target frames, following the same colored query points.
The selected video scenes do not appear in the main-paper figures.
These examples illustrate tracking across object motion, camera motion, and changes in appearance.

\FloatBarrier

{
    \small
    \bibliographystyle{ieeenat_fullname}
    \bibliography{main}
}

\end{document}